\documentclass{midl} 

\usepackage{mwe} 
\usepackage{soul}
\usepackage{parskip}
\usepackage{graphicx}
\usepackage{multirow}

\jmlrproceedings{MIDL}{Medical Imaging with Deep Learning}
\jmlrpages{}
\jmlryear{2026}

\jmlrworkshop{Short Paper Track}
\jmlrvolume{}
\editors{Accepted at MIDL 2026 -- Short Paper Track}

\title[TTE: Test-Time Explainability in Pretrained Black-Box CNNs]{TTE-CAM: Self-Explainable Class Activation Maps for Pretrained Black-Box CNNs}

\midlauthor{
\Name{Kerol Djoumessi\nametag{$^{1}$}} \orcid{0009-0004-1548-9758} \Email{kerol.djoumessi-donteu@uni-tuebingen.de}
\AND
\Name{Philipp Berens\nametag{$^{1,2}$}} \orcid{0000-0002-0199-4727} \Email{philipp.berens@uni-tuebingen.de}\\
\addr $^{1}$ Hertie Institute for AI in Brain Health, University of T\unexpanded{\"u}bingen, Germany \\
\addr $^{2}$ T\unexpanded{\"u}bingen AI Center, University of T\unexpanded{\"u}bingen, Germany
}

\begin{document}

\maketitle

\begin{abstract}
    Convolutional neural networks (CNNs) achieve state-of-the-art performance in medical image analysis yet remain opaque, limiting adoption in high-stakes clinical settings. Existing approaches face a fundamental trade-off: post-hoc methods provide unfaithful approximate explanations, while inherently interpretable architectures are faithful but often sacrifice predictive performance. We introduce TTE-CAM, a test-time framework that bridges this gap by converting pretrained black-box CNNs into self-explainable models via a convolution-based replacement of their classification head, initialized from the original weights. The resulting model preserves black-box predictive performance while delivering built-in faithful explanations competitive with post-hoc methods, both qualitatively and quantitatively. The code is available at \url{https://github.com/kdjoumessi/Test-Time-Explainability}.

\end{abstract}


\section{Introduction} 
    Convolutional neural networks (CNNs) achieve human-level performance across many tasks, including medical image analysis \cite{liu2019comparison}, yet their opaque decision processes limit interpretability and hinder adoption in high-stakes clinical settings \cite{ratti2022explainable}. 
    Existing explainability approaches face a fundamental trade-off: post-hoc methods generate saliency maps from the model that do not directly drive the output, making them inherently unfaithful \cite{adebayo2018sanity}. In contrast, interpretable-by-design architectures \cite{rudin2019stop, djoumessi2024actually} are faithful---their predictions are computed from the explanation---but they often require complex training or involve a trade-off in predictive performance. Bridging this gap by transforming high-performing black-box CNNs into self-explainable models without retraining or loss of accuracy remains an open challenge.

    We propose TTE-CAM, an architectural reformulation of class activation maps (CAMs) that transforms pretrained black-box CNNs into self-explainable models by replacing the classification head with $1 \times 1$ convolution layers initialized from the original weights. This reformulation yields built-in CAMs that serve as the sole input to the final prediction, enabling linearly interpretable decisions without post-hoc overhead.
    Unlike SoftCAM \cite{djoumessi2025soft}, which requires retraining, and conventional CAM-based post-hoc methods that derive explanations by weighting penultimate feature maps using different methods such as classification layer weights, gradients, or perturbations, 
    TTE-CAM integrates CAMs directly into the architecture. This preserves predictive performance while providing faithful, built-in explanations that are competitive with post-hoc approaches, as demonstrated on two medical imaging classification tasks.

\section{Materials and Methods}
    \begin{figure}[t]
        \centering
        \includegraphics[width=\textwidth]{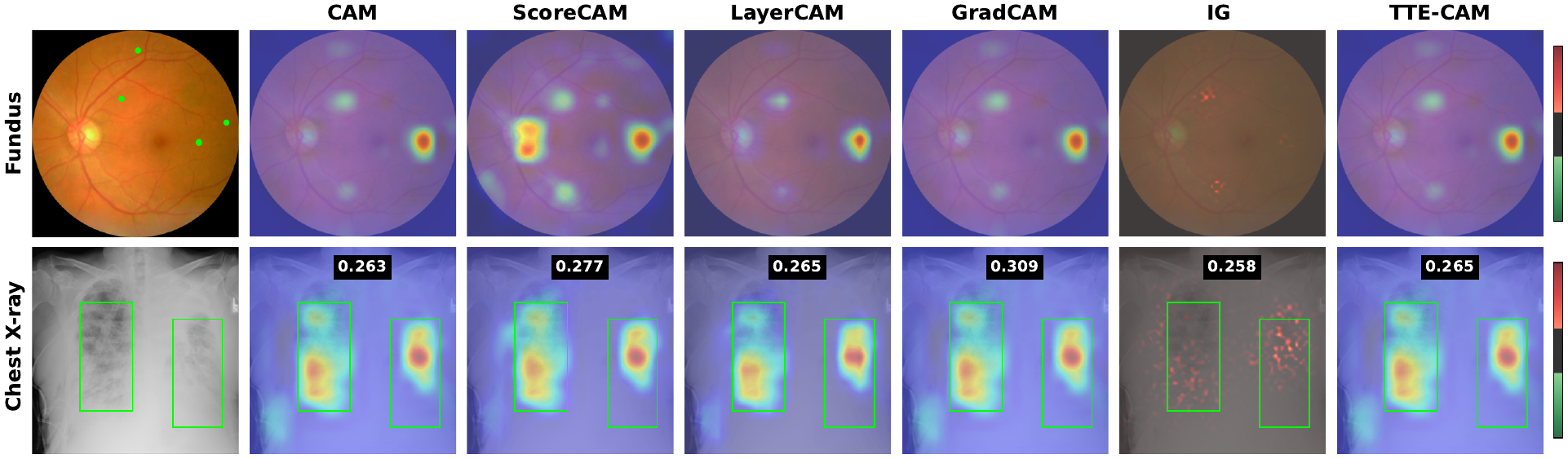}
        \caption{\textbf{Qualitative explanations comparison.} The first column shows a DR fundus with clinical annotations (green markers) and a pneumonia CXR with ground-truth bounding boxes (green). Columns 2-6 show post-hoc saliency maps; the last column shows TTE-CAM explanations. CXR scores indicate activation precision. Both samples were randomly selected from their corresponding disease classes.}
        \label{fig1}
    \end{figure}
    
    \paragraph{Datasets.} TTE-CAM was evaluated on two public medical imaging datasets. The Kaggle fundus Diabetic Retinopathy (DR) dataset \cite{kaggle_dr_detection} was used for binary classification of No DR (grade 0) versus DR (grades 1–4), while the RSNA Chest X-Ray (CXR) dataset \cite{rsna_dataset} was used for pneumonia detection. For explanation evaluation, RSNA bounding box annotations and clinical annotations from $65$ DR fundus images \cite{djoumessi2025inherently} were used for quantitative and qualitative assessment, respectively.

    \paragraph{Method.} 
    TTE-CAM reformulates the classification head of pretrained CNNs by removing the global average pooling (GAP) layer and replacing the fully connected layer (FCL) with a $1 \times 1$ convolutional layer comprising $C$ filters, where $C$ is the number of classes. Because a FCL is equivalent to a $1 \times 1$ convolution \cite{donteu2023sparse}, the pretrained classification weights can be transferred directly without retraining. This reformulation mirrors the original CAM architecture \cite{zhou2016learning}, in which class activation maps are obtained post-hoc by weighting feature maps with classification layer weights---here integrated into the architecture. The resulting layer produces built-in CAMs that are spatially averaged to compute class scores and then passed through a softmax to obtain the final predictions.   

    \paragraph{Post-hoc baseline.} TTE-CAM was compared against five post-hoc explainability methods from three families: gradient-free (CAM, ScoreCAM) \cite{zhou2016learning, wang2020score}, gradient-based (GradCAM, LayerCAM) \cite{selvaraju2017grad, jiang2021layercam}, and the backpropagation-based Integrated Gradients (IG) \cite{sundararajan2017axiomatic}. 
    
    \paragraph{Evaluation metrics.} Predictive performance was evaluated using accuracy (Acc.) and area under the curve (AUC). Explanation quality was assessed with three metrics ($k=10$): \emph{top-k sensitivity} \cite{yeh2019fidelity}, measuring the relative drop in predicted probability after masking the top-k most relevant regions; \emph{top-k localization}, quantifying the overlap between the top-k activated regions and annotated lesions; and \emph{activation precision} \cite{djoumessi2025soft}, measuring the fraction of activations within ground-truth bounding boxes. 

\section{Results}
    TTE-CAM was applied to a ResNet-50 \cite{he2016deep} trained on each dataset, with the checkpoint achieving the best validation accuracy used at test-time\footnote{The code is available at \url{https://github.com/kdjoumessi/Test-Time-Explainability}}. Replacing the FCL with a $1 \times 1$ convolutional classifier preserved predictive performance, yielding $\text{Acc.}=0.899$, $\text{AUC}=0.923$ for DR and $\text{Acc.}=0.953$, $\text{AUC}=0.988$ for pneumonia detection. 
    
    Qualitative (Fig.\,\ref{fig1}) and quantitative (Tab.\,\ref{tab1}) results show that TTE-CAM produces explanations similar to CAM and competitive with other methods across both datasets, while being built-in by design. The sparse fundus annotations are better suited for top-k localization, whereas the denser CXR bounding boxes are better suited for activation precision. 

    \begin{table}[]
        \small 
        \centering
        \begin{tabular}{c|l|c|c|c|c|c|c}
            \hline
             & \textbf{Metrics} & \textbf{CAM} & \textbf{S. CAM} & \textbf{L. CAM} & \textbf{G. CAM} & \textbf{IG} & \textbf{TTE-CAM} \\
            \hline
           \multirow{2}{*}{\rotatebox{0}{Fundus}} & Topk Prec. $\uparrow$ & $.33 \pm .29$ & $.28 \pm .22$ & $.31 \pm .26$ & $.39 \pm .28$ & $.39 \pm .28$ & $.33 \pm .28$ \\
              & Topk Sens. $\downarrow$ & $0.629$ & $0.668$ & $0.644$ & $0.629$ & $0.605$ & $0.629$  \\
           \hline
           \multirow{2}{*}{\rotatebox{0}{CXR}} & Acti. Prec. $\uparrow$ & $.12 \pm .09$ & $.13 \pm .10$ & $.12 \pm .09$ & $.13 \pm .10$ & $.12 \pm .09$ & $.12 \pm .09$  \\
              & Topk Sens. $\downarrow$ & $0.953$ & $0.959$ & $0.955$ & $0.953$ & $0.963$ & $0.953$  \\
           \hline
        \end{tabular}
        \caption{\textbf{Quantitative explanation comparison.} $\uparrow$ higher is better; $\downarrow$ lower is better.}
        \label{tab1}
    \end{table}

\section{Discussion and Conclusion}
    We show that pretrained back-box CNNs can provide built-in explanations at inference time by replacing the classification head with convolutional classifiers. TTE-CAM preserves the original predictive performance while producing explanations competitive with five post-hoc baselines spanning gradient-free, gradient-based, and backpropagation-based methods. Like post-hoc methods, it leverages pretrained weights without retraining, but generates explanations in a single forward pass, in contrast to post-hoc methods that require one forward pass per class (e.g. GradCAM) or multiple passes per class (e.g. ScoreCAM). 
    Importantly, the contribution of TTE-CAM is not improved localization performance over CAM-based methods, since the resulting explanations rely on similar feature weighting mechanisms, but the integration of explanations directly into the prediction pipeline. Unlike post-hoc attribution methods, TTE-CAM produces explanations directly through the forward computation, eliminating dependence on gradients, perturbations, hooks, or external explainability procedures. This makes explanations deterministic, reproducible, and available by design at inference time, while turning pretrained CNNs into lightweight interpretable-by-design models without retraining.
    
    TTE-CAM explanations are identical to CAM and competitive with other baselines, sharing a related feature map weighting mechanism. 
    While post-hoc explanations are often implemented as auxiliary analysis tools, TTE-CAM exposes the spatial evidence driving predictions as an explicit architectural component, which may simplify deployment and auditing in clinical settings where reliability and computational overhead are important considerations.
    Like all CAM-based methods, reliance on low-resolution feature maps can produce coarse explanations, limiting fine-grained localization, as observed in DR. 
    Weight transfer constraints further restrict applicability to architectures where the final feature map channel dimension matches the classifier input size (e.g., ResNet and DenseNet), excluding models such as VGG. 
    Future work could address these constraints, extend this mechanism to vision transformers for built-in attention map explanations \cite{djoumessi2025hybrid}, and explore explanation-aware workflows such as uncertainty estimation, clinical verification, or test-time intervention.

\midlacknowledgments{This project was supported by the Hertie, the German Science Foundation (Excellence Cluster EXC 2064 ``Machine Learning—New Perspectives for Science'', project number 390727645; BE 5601/14-1, project number 571331899).
}

\bibliography{midl-samplebibliography}

\end{document}